# Deep learning is effective for the classification of OCT images of normal versus Age-related Macular Degeneration

Cecilia S. Lee MD[1], Doug M. Baughman BS[1], Aaron Y. Lee MD MSCI[1]

[1] Department of Ophthalmology
	University of Washington School of Medicine, Seattle WA

Corresponding Author:
	Aaron Y. Lee
	Assistant Professor
	Department of Ophthalmology
	University of Washington
	Box 359608, 325 Ninth Avenue
	Seattle WA 98104
	Ph: (206) 543-7250
	Email: leeay@uw.edu

Financial support: National Eye Institute, Bethesda, Maryland (grant no.: K23EY02492 [C.S.L.]); Latham Vision Science Innovation Grant, Seattle WA (C.S.L., D.M.B.)  and Research to Prevent Blindness, Inc., New York, New York (C.S.L., A.Y.L.).  The sponsors and funding organizations had no role in the design or conduct of this research.

Conflict of interest: No conflicting relationship exists for any author.

Running head: Deep learning for classification of macular OCT images

**Abbreviations and acronyms**
AMD: Age-related Macular Degeneration
AUROC: Area Under the Receiver Operator Curve
CAD: Computer Aided Diagnosis
EMR: Electronic Medical Records
FDA: Food and Drug Administration
GPU: Graphics Processing Unit
ICD-9: International Classification of Diseases, 9th edition
OCT: Optical Coherence Tomography
ReLU: Rectified Linear Unit
ROC: Receiver Operator Curve
RPE: Retinal Pigmented Epithelium
SD: Standard Deviation
UW: University of Washington

**ABSTRACT**

Objective: The advent of Electronic Medical Records (EMR) with large electronic imaging databases along with advances in deep neural networks with machine learning has provided a unique opportunity to achieve milestones in automated image analysis. Optical coherence tomography (OCT) is the most commonly obtained imaging modality in ophthalmology and represents a dense and rich dataset when combined with labels derived from the EMR. We sought to determine if deep learning could be utilized to distinguish normal OCT images from images from patients with Age-related Macular Degeneration (AMD).

Design: EMR and OCT database study

Subjects: Normal and AMD patients who had a macular OCT.

Methods: Automated extraction of an OCT imaging database was performed and linked to clinical endpoints from the EMR. OCT macula scans were obtained by Heidelberg Spectralis, and each OCT scan was linked to EMR clinical endpoints extracted from EPIC. The central 11 images were selected from each OCT scan of two cohorts of patients: normal and AMD. Cross-validation was performed using a random subset of patients. Receiver operator curves (ROC) were constructed at an independent image level, macular OCT level, and patient level.

Main outcome measure: Area under the ROC.

Results: Of a recent extraction of 2.6 million OCT images linked to clinical datapoints from the EMR, 52,690 normal macular OCT images and 48,312 AMD macular OCT images were selected. A deep neural network was trained to categorize images as either normal or AMD. At the image level, we achieved an area under the ROC of 92.78% with an accuracy of 87.63%. At the macula level, we achieved an area under the ROC of 93.83% with an accuracy of 88.98%. At a patient level, we achieved an area under the ROC of 97.45% with an accuracy of 93.45%. Peak sensitivity and specificity with optimal cutoffs were 92.64% and 93.69% respectively.

Conclusions: Deep learning techniques achieve high accuracy and is effective as a new image classification technique. These findings have important implications in utilizing OCT in automated screening and the development of computer aided diagnosis tools in the future.

**INTRODUCTION**

Optical coherence tomography (OCT) has become the most commonly used imaging modality in ophthalmology with 4.39 million, 4.93 million, and 5.35 million OCTs performed in 2012, 2013, and 2014 respectively in the US Medicare population.[1] Since its development in 1991,[2] a 70-fold increase in OCT use for diagnosing age-related macular degeneration (AMD) was reported between 2002 and 2009.[3] Furthermore, since the development of anti-angiogenic agents, OCT has become a critical tool for baseline retinal evaluation prior to initiation of therapy and monitoring therapeutic effect.[4,5] This increase in the use of OCT imaging, with images stored in large electronic databases, highlights the ever-increasing time and effort spent by providers interpreting images.

The key OCT findings in AMD, including drusen, retinal pigmented epithelium (RPE) changes, and subretinal and intraretinal fluid,[4] share some common OCT features that are distinctively different from a normal retina.[6] Correct identification of these characteristics allows for precise management of neovascular AMD and guides the decision of whether intravitreal therapy with anti-VEGF agents should be given or not.[7–9] Computer aided diagnosis (CAD) has the potential for allowing more efficient identification of pathological OCT images and directing the attention of the clinician to regions of interest on the OCT images.

The concept of CAD is not novel and has been applied in radiology, a field where the increasing demand of imaging studies has begun to outpace the capacity of practicing radiologists.[10] A number of CAD systems have been approved by the Food and Drug Administration (FDA) for lesion detection and volumetric analysis in mammography, chest radiography, and chest computed tomography.[11]

Traditional image analysis required the manual development of convolutional matrices applied to an image for edge detection and feature extraction. In addition, prior work on OCT image classification of diseases has relied on machine learning techniques such as Principal Components Analysis, Support Vector Machine, or Random Forest.[12–14] However recently, there has been a revolutionary step forward in machine learning techniques with the advent of deep learning where a many-layered neural network is trained to develop these convolutional matrices purely from training data.[15] Specifically, the development of convolutional neural network layers allowed for significant gains in the ability to classify images and detect objects in a picture.[16–18] Within ophthalmology, deep learning has been recently applied at a limited capacity to automated detection of diabetic retinopathy from fundus photos, visual field perimetry in glaucoma patients, grading of nuclear cataracts, and segmentation of foveal microvasculature, each with promising initial findings.[19–22]

While deep learning has revolutionized the field of computer vision, their application is usually limited due to the lack of large training sets. Often several tens of thousands of examples are required before deep learning can be used effectively. With the ever increasing use of OCT as an imaging modality in ophthalmology along with the use of codified structured clinical data in the Electronic Medical Record (EMR), we sought to link two large datasets together to use as a training set for developing a deep learning algorithm to distinguish AMD from normal OCT images.

**METHODS**

This study was approved by the Institutional Review Board of the University of Washington (UW) and was in adherence with the tenets of the Declaration of Helsinki and the Health Insurance Portability and Accountability Act.

*OCT and EMR Extraction*
Macular OCT scan were extracted using an automated extraction tool from the Heidelberg Spectralis imaging database from 2006 to 2016. Each macular scan was obtained using a 61 line raster macula scan, and every image of each macular OCT was extracted. The images were then linked by patient medical record number and dates to the clinical data stored in EPIC. Specifically, all clinical diagnoses and the dates of every clinical encounter, macular laser procedures, and intravitreal injections were extracted from the EPIC Clarity tables.

*Patient and Image Selection*
A normal patient was defined as having no retinal ICD-9 diagnosis and better than 20/30 vision in both eyes during the entirety of recorded clinical history at UW. An AMD patient was defined as having the ICD-9 diagnosis of AMD (362.50, 362.51, and 362.52) by a retina specialist, at least one intravitreal injection in either eye, and worse than 20/30 vision in the better seeing eye. Patients with other macular pathology by ICD-9 code were excluded. These parameters were chosen *a priori* to ensure that macular pathology was most likely present in both eyes in the AMD patients and absent in both eyes in the normal patients. Consecutive images of patients meeting this criteria were included and no images were excluded due to image quality. Labels from the EMR were then linked to the OCT macular images, and the data was stripped of all protected health identifiers.

As most of the macular pathology is concentrated in foveal region, the decision was made a priori to select the central 11 images from each macular OCT set, and each image was then treated independently, labeled as either normal or AMD. The images were histogram equalized and the resolution down-sampled to 192x124 due to limitations of memory. The image set was then divided into two sets with 20% of the patients in each group placed into the validation set and the rest were used for training. Care was taken to ensure that the validation set and the training set contained images from mutually exclusive group of patients (i.e. no single patient contributed images to both the training and validation set). The order of images was then randomized in the training set.

*Deep Learning Classification Model*

A modified version of the VGG16 convolutional neural network[23] was used as the deep learning model for classification (Figure 1). Weights were initialized using the Xavier algorithm.[24] Training was then performed using multiple iterations each with a batch size of 100 images with a starting learning rate of 0.001 with stochastic gradient descent optimization. At each iteration, the loss of the model was recorded, and at every 500 iterations, the performance of the neural network was assessed using cross-validation with the validation set. The training was stopped when the loss of the model decreased and the accuracy of the validation set decreased.

An occlusion test[17] was performed to identify the areas most contributing to the neural network assigning the category of AMD. A blank 20x20 pixel box was systematically moved across every possible position in the image and the probabilities were recorded. The highest drop in the probability represents the region of interest that contributed the highest importance to the deep learning algorithm.

Caffe (http://caffe.berkeleyvision.org/) and Python (http://www.python.org) were used to perform deep learning. All training occurred using the NVIDIA Pascal Titan X Graphics Processing Unit (GPU) with NVIDA cuda (v8.0) and cu-dnn (v5.5.1) libraries (http://www.nvidia.com). Macular OCT level analysis was performed by averaging the probabilities of the images obtained from the same macular OCT. Patient level analysis was performed by averaging the probabilities of the images obtained from the same patient. Receiver-operator curves (ROC) were constructed using the probability output from the deep learning model. Statistics were performed using R (http://www.r-project.org).

**RESULTS**

We successfully extracted 2.6 million OCT images of 43,328 macular OCT scans from 9,285 patients. After linking the macular OCT scans to the EMR, 48,312 images from 4,392 normal OCT scans and 52,690 images from 4,790 AMD OCT scans were selected. A total of 80,839 images (41,074 from AMD, 39,765 from normal) were used for training and 20,163 images (11,616 from AMD, 8,547 from normal) were used for validation.

After 8,000 iterations of training the deep learning model, the training was stopped due to overfitting occurring after that point.(Figure 2) ROC curves are shown at the image level, OCT macular level, and the patient level in Figure 3. The average time to evaluate a single image after training was complete was 4.97 milliseconds.

At the level of each individual image, we achieved an accuracy of 87.63% with a sensitivity of 84.63% and a specificity of 91.54%. After constructing an ROC curve, the peak sensitivity and specificity with optimal cutoffs were 87.08% and 87.05% respectively. The area under the ROC curve (AUROC) was 92.77%.

By grouping the images in the same OCT scan and averaging the probabilities from each image, we achieved an accuracy of 88.98% with a sensitivity of 85.41% and a specificity of 93.82%. After constructing the ROC curve, the peak sensitivity and specificity with optimal cutoffs were 88.63% and 87.77%. The AUROC was 93.82%.

By averaging the probabilities from each image from the same patient, we achieved an accuracy of 93.45% with a sensitivity of 83.82% and a specificity of 96.40%. After constructing the ROC curve, the peak sensitivity and specificity with optimal cutoffs were 92.64% and 93.69%. The AUROC was 97.46%.

Example images from the occlusion test are shown in Figure 4 showing that the neural network was successfully able to identify pathological regions on the OCT. These areas represent the most critical area in each image to the trained network in categorizing the image as AMD.

**DISCUSSION**

The ever increasing use of digital imaging and EMRs provide opportunities to create deep and rich datasets for analysis. Our study demonstrates that a deep learning neural network was effective at distinguishing AMD from normal OCT images, and its accuracy was even higher when aggregate probabilities at the OCT macular scan and patient level were combined. The increase in the AUROC mainly occurred with an increase in the sensitivity (Figure 3) with inclusion of more images when aggregated. This most likely occurred as AMD infrequently affects the entire macula and normal appearing OCT images may be mixed in with a macular OCT scan obtained from an AMD patient. Another possible explanation is that the etiology of the CNVM may be incorrect in the EMR. For example, our AMD labeled images may have included myopic CNVM which has localized pathology with relatively normal OCTs outside the CNVM area. Thus, adding additional images may explain the increased sensitivity.

Only limited applications of deep learning models exist in ophthalmology. Abramoff et al. reported sensitivity of 96.8% and specificity of 87.0% in detecting referable diabetic retinopathy in 874 patients using deep learning model, but no details of the algorithm were included.[20] Asaoka et al. used a deep learning method to differentiate the visual fields of 27 patients with preperimetric open angle glaucoma from 65 controls.[19] The AUROC of 92.6% was achieved with this study's feed-forward network classifier, but the algorithm only used four layers of neurons. In contrast, our study deep learning algorithm included 21 layers of neurons with state of the art convolutional neural networking layers. Other smaller studies that have applied automated OCT classification algorithms did not integrate deep learning strategies and was limited by small sample size ranging from 32 to 384.[12,19,20]

Our deep learning algorithm is a novel application to OCT classification in ophthalmology. To our best knowledge, the use of training and validation images from large EMR extraction has never been shown. In order to verify how the deep learning algorithm categorized the images as AMD, we performed an occlusion test where we systematically occluded every location in the image with a blank 20x20 pixel area. The deep learning neural network successfully identified key areas of interest on the OCT image, which corresponded to the areas of pathology (Figure 4).

The application of occlusion testing provides insight into the trained deep learning model and which features were most important in distinguishing AMD images from normal images. In Figure 4C, occlusion testing did not show high intensity dependence in the area of nasal high choroidal transmission suggesting that the classifier was not using this as an important feature. One possible explanation is that in distinguishing normal from AMD OCT images, the classifier already achieved very low loss (Figure 2B) using the identified features and never discovered further improvements. Further studies could be performed where a classifier is specifically trained to distinguish specific AMD OCT features such as drusen, subretinal fluid, pigment epithelial detachments, and high choroidal transmission from each other.

Our study findings have several limitations. We only included images from patients who met our study criteria and the neural network was only trained on these images. However, we included consecutive real-world images and did not exclude images with poor quality. In addition, the training of this model was only done on images from a single academic center and the external generalizability is unknown. Future studies would include expanding the number of diagnoses, using all images from a macular OCT scan, including images from different OCT manufacturers, and validation on OCT scans from other institutions.

In the future, our approach may be used to develop deep learning models that can have a number of wide-reaching applications. First, the models can be applied to various retinal or choroidal pathologies in which OCT evaluations are essential, including diabetic retinopathy or retinal vein occlusions. Second, in future studies using deep learning, automated macular OCT classification could be used as a screening tool for retinal pathology when the hardware cost of OCT machines decreases. This automated classification could be added to the majority of OCT machines being used in clinical practice without the need of integrating GPU as the inference step of deep learning is computationally inexpensive compared to training and can be run on standard computers. The automated classification feature will likely be beneficial in a large screening model such as in an AMD screening system. Finally, deep learning models can identify concerning macular OCT images and efficiently display them to the clinician to aid in the diagnosis and treatment of macular pathology, much like computer-aided diagnosis models found in radiology.

In conclusion, we demonstrate the ability of a deep learning model to distinguish AMD versus normal OCT images and show encouraging results for the first application of deep learning to OCT images. Future follow up studies will include widening the number of diseases and showing external validity of the model using images from other institutions.

ACKNOWLEDGEMENTS

We would like to thank Dr. Michael Boland for his assistance in identifying EPIC Clarity database values, the UW eScience Institute for their infrastructure support, and especially NVIDIA Corporation for their hardware GPU donation.

FIGURE LEGENDS

FIGURE 1: Schematic of the deep learning model used. A total of 21 layers with Rectified Linear Unit (ReLU) activations were used.

FIGURE 2: Learning curve of the training of the neural network with accuracy (A) and loss (B). Each iteration represents 100 images being trained on the neural network.

FIGURE 3: Receiver-operator curves of three levels of classification. Image level classification was performed by considering each image independently. Macula level classification was performed by averaging the probabilities of all the images in a single macular volume. Patient level classification was performed by averaging the probabilities of all the images belonging to a single patient.

FIGURE 4: Examples of identification of pathology by deep learning algorithm. Images of optical coherence tomography (OCT) with age-related macular degeneration (AMD) pathology (A, B, C) are used as input images and hotspots (D, E, F) are identified using an occlusion test from the deep learning algorithm. The intensity of the color is determined by the drop in the probability of being labeled AMD when occluded.

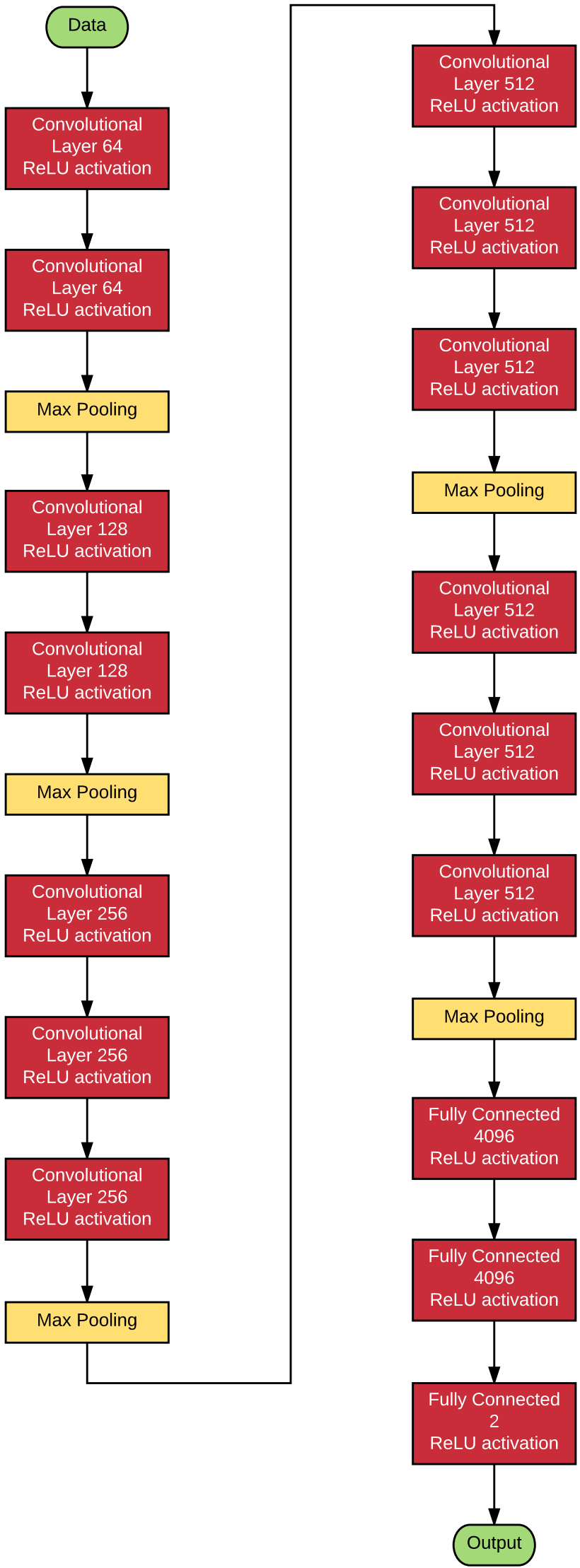

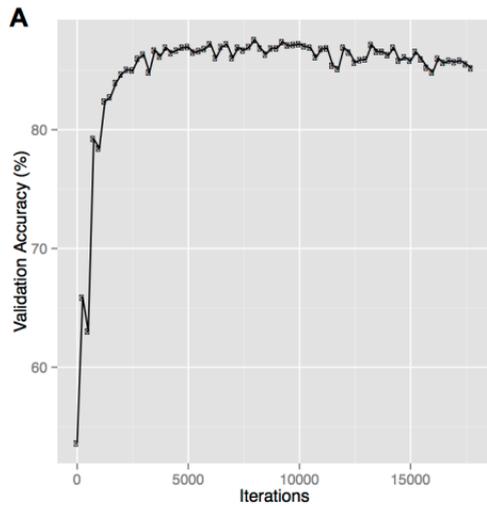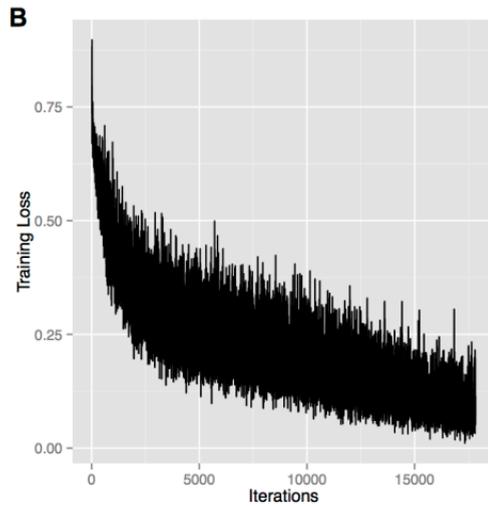

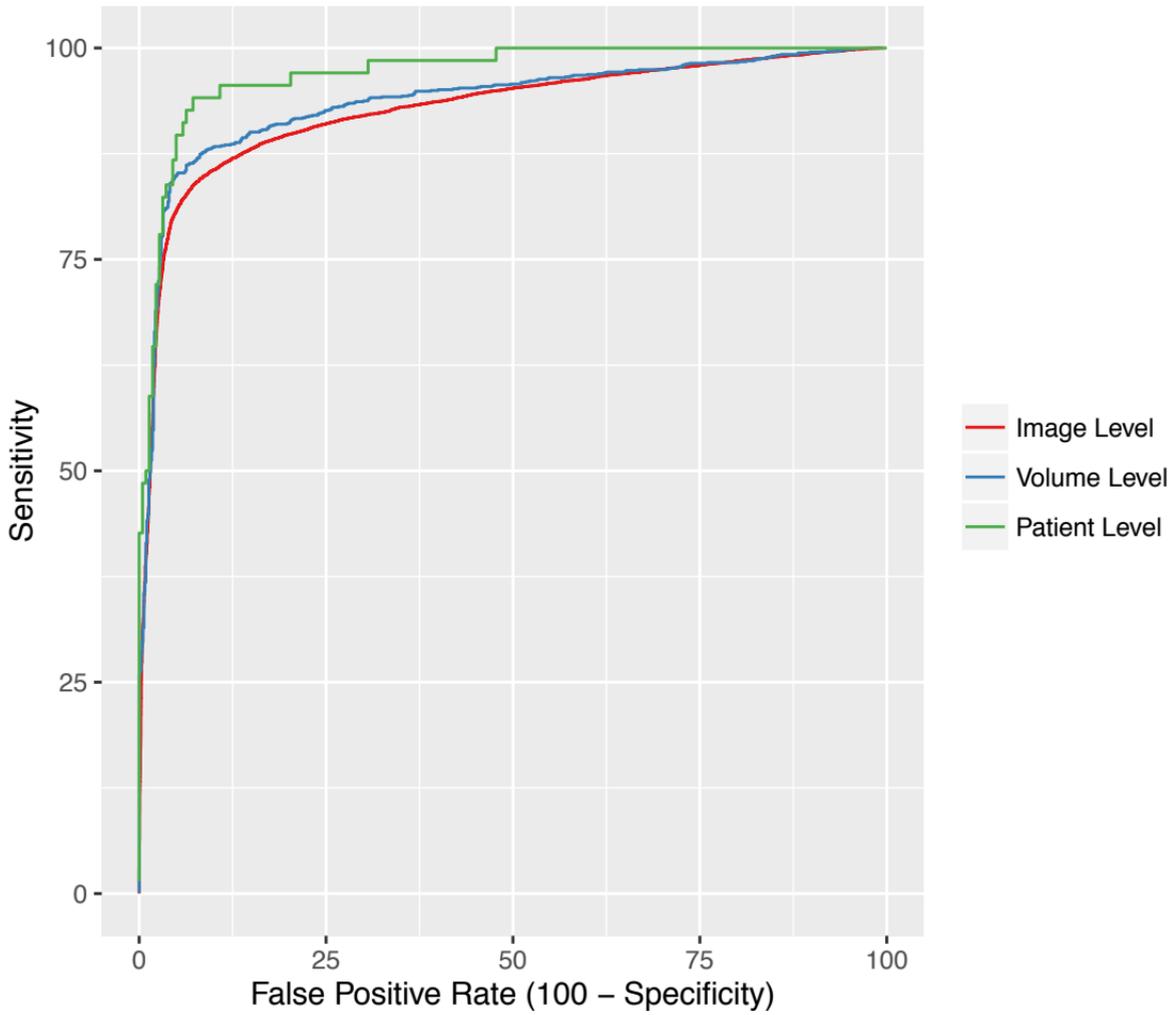

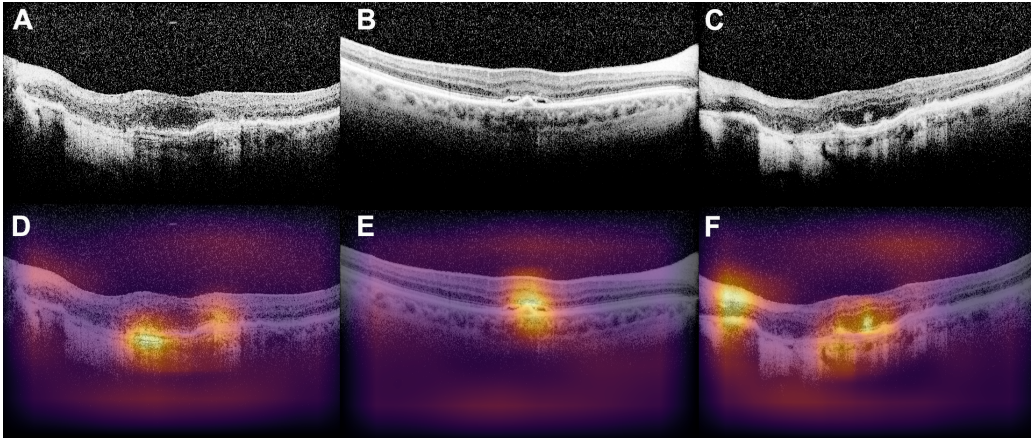